# Random Transformation of Image Brightness for Adversarial Attack

Bo Yang, Kaiyong Xu, Hengjun Wang, Hengwei Zhang

**Abstract:** Deep neural networks are vulnerable to adversarial examples, which are crafted by adding small, human-imperceptible perturbations to the original images, but make the model output inaccurate predictions. Before deep neural networks are deployed, adversarial attacks can thus be an important method to evaluate and select robust models in safety-critical applications. However, under the challenging black-box setting, the attack success rate, *i.e.*, the transferability of adversarial examples, still needs to be improved. Based on image augmentation methods, we found that random transformation of image brightness can eliminate overfitting in the generation of adversarial examples and improve their transferability. To this end, we propose an adversarial example generation method based on this phenomenon, which can be integrated with Fast Gradient Sign Method (FGSM)-related methods to build a more robust gradient-based attack and generate adversarial examples with better transferability. Extensive experiments on the ImageNet dataset demonstrate the method's effectiveness. Whether on normally or adversarially trained networks, our method has a higher success rate for black-box attacks than other attack methods based on data augmentation. We hope that this method can help to evaluate and improve the robustness of models.

## 1. Introduction

In image recognition, some experiments on standard test sets have proven that deep neural networks (DNNs) have higher recognition ability than that of humans [1-4]. However, while deep learning brings great convenience, it also brings security problems. For an abnormal input, the question remains whether DNNs can obtain satisfactory results. DNNs have been shown to be highly vulnerable to attacks from adversarial examples [5,6], because to add perturbations to an original input image that are imperceptible to humans will cause misclassification of the models. Adversarial examples normally have a certain degree of transferability, *i.e.*, those generated for one model may also be adversarial to another, which enables black-box attacks [7]. Adversarial examples with strong attack performance can thus be used as an important tool to evaluate and improve the robustness of models.

Although adversarial examples are generally transferable, to further improve their transferability for effective black-box attacks remains to be explored. In the search of more transferable adversarial examples, some gradient-based attacks have been proposed, such as single-step [6] and iterative [8,9] methods. These show powerful attack capabilities in the white-box setting, but their success rates are relatively low in the black-box setting, which we attribute to overfitting of adversarial examples, *i.e.*, the difference in attack ability of an adversarial example under white-box and black-box settings is similar to that of the same neural network on training and test sets. Therefore, we can apply methods that improve the performance of deep learning models to the generation of adversarial examples to eliminate overfitting and improve their transferability. Many methods have been proposed to improve DNN performance [1,2,9-12]; one of the most important is data augmentation [1,2], and it can prevent overfitting during training and improve the generalization ability of models.

We optimize the generation of adversarial examples based on data augmentation and propose the Random Transformation of Image Brightness Attack Method (RTM) to improve their transferability.

1) Inspired by data augmentation [1,2], we adapt the random transformation of image brightness to adversarial attacks, so as to effectively eliminate overfitting in the generation of adversarial examples and improve their transferability.

2) Our method is readily combined with gradient-based attack methods (*e.g.*, momentum iterative gradient-based [9] and diverse input [14] methods) to further boost the success rate of adversarial examples for black-box attacks.

Extensive experiments on the ImageNet dataset [13] have indicated that, compared to current data augmentation attack methods [14], our method, RT-MI-FGSM (Random Transformation of Image Brightness Momentum Iterative Fast Gradient Sign Method), has a higher success rate for black-box attacks in normally and adversarially trained models. Integrated with the diverse input method (DIM) [14], the resulting RT-DIM (Random Transformation of image brightness with Diverse

Input Method) can greatly improve the average attack success rate on adversarially trained models in black-box settings. The method of attacking ensemble models simultaneously is used to further improve the transferability of adversarial examples [7]. Under the ensemble attack, RT-DIM reaches an average success rate of 72.1% for black-box attacks on adversarially trained networks, which outperforms DIM by a large margin of 23.5%. We hope that the proposed attack method can help evaluate the robustness of models and effectiveness of defense methods.

## 2. Related Work

*2.1 Adversarial example generation*

Biggio et al. [15] presented a simple but effective gradient-based method that can be used to systematically assess the security of several widely-used classification algorithms against evasion attacks, indicating that traditional machine learning algorithms are vulnerable to adversarial examples. Szegedy et al. [5] reported the intriguing property that DNNs are fragile to adversarial examples and proposed the L-BFGS method to generate them. Goodfellow et al. [6] demonstrated the fast gradient sign method that can generate adversarial examples with one gradient step. It reduces the computation needed to generate adversarial examples and forms the basis of subsequent FGSM-related methods. Alexey et al. [8] extended FGSM to an iterative version, which greatly improved the success rate for white-box attacks and proved that adversarial examples also exist in the physical world. Dong et al. [9] proposed momentum-based iterative FGSM, improving the transferability of adversarial examples. Xie et al. [14] randomly transformed the original input images in each iteration to reduce overfitting and improve the transferability of adversarial examples. Dong et al. [16] used a set of translated images to optimize adversarial perturbations. To reduce computation, the gradient was calculated by convolving the gradient of the untranslated images with the kernel matrix, which can generate adversarial examples with better transferability. In addition, the fact that adversarial examples may exist in the physical world brings security threats to the practical application of DNNs [8,17].

*2.2 Defense methods against adversarial examples*

Many defense methods against adversarial examples have been proposed to protect deep learning models [18-25]. Adversarial training [6,26,27] is one of the most effective ways to improve the robustness of models, by injecting adversarial examples into training data. Xie et al. [20] found that the effectiveness of adversarial examples can be reduced through random transformation. Guo et al. [21] found a range of image transformations with the potential to remove adversarial perturbations while preserving the key visual information of an image. Samangouei et al. [22] used a generative model to purify adversarial examples by moving them back toward the distribution of the original clean image, thereby reducing their impact. Liu et al. [23] proposed a JPEG-based defensive compression framework that can rectify adversarial examples without affecting classification accuracy on benign data, alleviating the adversarial effect. Cohen et al. [25] proposed a randomized smoothing technique to obtain an ImageNet classifier with certified adversarial robustness. Tramèr et al. [27] proposed ensemble adversarial training, utilizing adversarial examples generated for other models to increase training data and further improve the robustness of models.

## 3. Methodology

Let $x$ be the original input image, $y$ the corresponding true label, and $\theta$ the parameter of the model. $J(\theta, x, y)$ is the loss function of the neural network, which is usually cross-entropy loss. We aim to generate an adversarial example $x^{adv}$ that is visually indistinguishable from $x$ by maximizing $J(\theta, x, y)$ to fool the model; *i.e.*, the model misclassifies the adversarial example $x^{adv}$. We use the $L_\infty$ norm bound, $\|x^{adv} - x\|_\infty \leq \varepsilon$, to limit adversarial perturbations. Hence adversarial example generation can be transformed to a condition constrained optimization problem:

$$\arg\max_{x^{adv}} J(\theta, x^{adv}, y), \quad s.t. \|x^{adv} - x\|_\infty \leq \varepsilon. \tag{1}$$

## 3.1 Gradient-based adversarial attack methods

We introduce several methods to generate adversarial examples.

*Fast Gradient Sign Method (FGSM).* FGSM [6] is one of the most basic methods, which searches adversarial examples in the direction of the loss gradient $\nabla_x J(\theta, x, y)$ with respect to the input and imposes infinity norm restrictions on adversarial perturbations. The updated equation is

$$x^{adv} = x + \varepsilon \cdot sign(\nabla_x J(\theta, x, y)) . \tag{2}$$

*Iterative Fast Gradient Sign Method (I-FGSM).* Kurakin et al. [8] proposed an iterative version of FGSM. It divides the gradient operation in FGSM into multiple iterations to eliminate the under-fitting caused by single-step attacks. It can be expressed as

$$x_0^{adv} = x , \quad x_{t+1}^{adv} = \text{Clip}_x^{\varepsilon} \{ x_t^{adv} + \alpha \cdot sign(\nabla_x J(\theta, x_t^{adv}, y)) \} , \tag{3}$$

where $\alpha$ is the step size of each iteration and $\alpha = \varepsilon / T$, where $T$ is the number of iterations. The Clip function restricts the adversarial example to be within the $\varepsilon$-ball of the original image $x$ to meet the infinity norm constraint. Experiments have shown that I-FGSM has a higher success rate for white-box attacks than FGSM, but with poorer transferability.

*Momentum Iterative Fast Gradient Sign Method (MI-FGSM).* MI-FGSM [9] is the first method to apply momentum to adversarial example generation, which can stabilize gradient update directions, improve convergence, and greatly increase the attack success rate. MI-FGSM differs from I-FGSM in the update directions of adversarial examples

$$x_0^{adv} = x , \quad g_0 = 0 , \quad g_{t+1} = \mu \cdot g_t + \frac{\nabla_x J(\theta, x_t^{adv}, y)}{\| \nabla_x J(\theta, x_t^{adv}, y) \|_1} , \tag{4}$$

$$x_{t+1}^{adv} = \text{Clip}_x^{\varepsilon} \{ x_t^{adv} + \alpha \cdot sign(g_{t+1}) \} , \tag{5}$$

where $\mu$ is the decay factor of the momentum term, and $g_t$ is the gradient weighted accumulation of the previous $t$ iterations.

*Diverse Input Method (DIM).* DIM [14] randomly transforms the original input with a given probability in each iteration to reduce overfitting. Transformations include random resizing and padding. This method is readily combined with other baseline attack methods to generate adversarial examples with better transferability. The random transformation equation is

$$T(X_t^{adv}; p) = \begin{cases} T(X_t^{adv}), & \text{with probability } p \\ X_t^{adv}, & \text{with probability } 1-p \end{cases} . \tag{6}$$

*Projected Gradient Descent (PGD).* PGD [18] is a strong iterative version of FGSM, which improves the attack success rate of adversarial examples.

## 3.2 Random transformation of image brightness attack method

Data augmentation [1,2] has been proven effective to prevent network overfitting during DNN training. Based on this, we propose the Random Transformation of Image Brightness Attack Method (RTM), which randomly transforms the brightness of the original input image with probability $p$ in each iteration to alleviate overfitting. It optimizes the adversarial perturbations of the image with randomly transformed brightness:

$$\arg\max_{x^{adv}} J(\theta, RT(x^{adv}; p), y) , \quad \text{s.t.} \| x^{adv} - x \|_{\infty} \leq \varepsilon , \tag{7}$$

$$RT(X^{adv}; p) = \begin{cases} RT(X^{adv}), & \text{with probability } p \\ X^{adv}, & \text{with probability } 1-p \end{cases} . \tag{8}$$

The random transformation function $RT(\cdot)$ randomly decreases the brightness of the input image at a random rate. The transformation probability $p$ controls the balance between the original input image and the transformed image. With this method, we can achieve effective attacks on the model through data augmentation, avoid overfitting attacks of white-box models, and improve the transferability of adversarial examples.

*3.3 Attack algorithms*

For the gradient processing of generating adversarial examples, RTM introduces data augmentation to alleviate overfitting. RTM is easily combined with MI-FGSM to form a stronger attack, which we refer to as RT-MI-FGSM (Random Transformation of image brightness Momentum Iterative Fast Gradient Sign Method). Our algorithm can be associated with the family of FGSM by adjusting its parameter settings. For example, RT-MI-FGSM degrades to MI-FGSM if $p=0$, *i.e.*, we can remove step 4 of algorithm 1 to realize MI-FGSM. Algorithm 1 summarizes the RT-MI-FGSM attack algorithm.

---

**Input:** A clean example $x$ with ground-truth label $y$; a classifier $f$ with loss function $J$;
**Input:** Perturbation size $\varepsilon$; maximum iterations $T$ and decay factor $\mu$.
**Output:** An adversarial example $x^{adv}$
1: $\alpha = \varepsilon / T$
2: $x_0^{adv} = x$; $g_0 = 0$
3: **for** $t = 0$ to $T-1$ **do**
4: $\quad$ Get $x_t^{adv}$ by $x_t^{adv} = RT(x_t^{adv}; p)$ ▷ apply random transformation of the input's brightness with the probability $p$
5: $\quad$ Get the gradients by $\nabla_x J(\theta, x_t^{adv}, y)$
6: $\quad$ Update $g_{t+1}$ by $g_{t+1} = \mu \cdot g_t + \dfrac{\nabla_x J(\theta, x_t^{adv}, y)}{\|\nabla_x J(\theta, x_t^{adv}, y)\|_1}$
7: $\quad$ Update $x_{t+1}^{adv}$ by Eq. (5)
8: **return** $x^{adv} = x_T^{adv}$

ALGORITHM 1: the details of RT-MI-FGSM

---

In addition, RTM can be combined with DIM to form RT-DIM, further improving the transferability of adversarial examples. The algorithm of this attack method is shown in Appendix A.

## 4. Experiments

We conduct experiments to evaluate our method's effectiveness. Below, we specify the experimental settings, show the results of attacking a single network, validate our method on ensemble models, and discuss the hyper-parameters that affect the results.

*4.1 Experimental setup*

*Dataset.* It is less meaningful to generate adversarial examples from the original images that are already classified wrongly. We randomly selected 1000 images belonging to 1000 categories (i.e., one image per category) from the ImageNet verification set, which were correctly classified by our testing networks. All images were adjusted to $299 \times 299 \times 3$.

*Models.* We consider seven networks. The four normally trained networks are Inception-v3 (Inc-v3) [28], Inception-v4 (Inc-v4) [29], Inception-Resnet-v2 (IncRes-v2) [29], and Resnet-v2-101 (Res-101) [30]; the three adversarially trained networks [27] are ens3-adv-Inception-v3 (Inc-v3$_{ens3}$), ens4-adv-Inception-v3 (Inc-v3$_{ens4}$), and ens-adv-Inception-ResNet-v2 (IncRes-v2$_{ens}$).

*Baselines.* We integrated our method with MI-FGSM [9] and DIM [14] to evaluate the improvement of RTM over these baseline methods.

*Implementation details.* For the hyper-parameters, we follow the default settings in [9] with the maximum perturbation $\varepsilon = 16$, number of iterations $T = 10$, and step size $\alpha = 1.6$. For MI-FGSM, the decay factor is defaulted to $\mu = 1.0$. For

DIM, we adopt the default settings. For our methods, $p$ is set to 0.5 for the random transformation function $RT(X;p)$, and to 1.0 when RTM is combined with DIM. For transformation operations $RT(\cdot)$, the brightness of the input image $x$ is randomly adjusted to $B*r$, where $B$ is the original brightness of the input image and $r \in (1/16, 1]$ is the adjustment rate. For intuitive understanding, Appendix B shows some images after random brightness transformation.

*4.2 Attacking a single network*

We first perform adversarial attacks on a single network. We use I-FGSM, MI-FGSM, DIM, and RT-MI-FGSM to generate adversarial examples only on the normally trained networks and tested them on all seven networks. The results are shown in Table 1, where the success rate is the model classification error rate with adversarial examples as input. We also combine RTM and DIM as RT-DIM. $p$ is set to 1.0 for RTM in this case. The test results on the seven networks are shown in Table 2.

The results in Table 1 show that the attack success rates of RT-MI-FGSM under mostly black-box settings are much higher than those of other baseline attacks. It also has higher attack success rates than the DIM attack method based on data augmentation, and maintains relatively high white-box attack success rates. For example, when generating adversarial examples on the Inc-v3 network to attack the Inc-v4 network, the success rate for black-box attacks of RT-MI-FGSM reaches 71.4%, the highest among these methods. RT-MI-FGSM also performs better on the adversarially trained networks. Compared to the other three attack methods, our method greatly improves the success rates for black-box attacks. For example, when generating adversarial examples on the Inc-v3 network to attack the adversarially trained networks, the average attack success rates of RT-MI-FGSM and MI-FGSM are 24.6% and 12.2%, respectively. This 12.4% enhancement demonstrates that our method can effectively improve the transferability of adversarial examples.

TABLE 1: The success rates (%) of adversarial attacks against seven models under single model setting. Adversarial examples are crafted on Inc-v3, Inc-v4, IncRes-v2, and Res-101, respectively, using I-FGSM, MI-FGSM, DIM, and RT-MI-FGSM. * indicates white-box attacks.

| Model | Attack | Inc-v3 | Inc-v4 | IncRes-v2 | Res-101 | Inc-v3$_{ens3}$ | Inc-v3$_{ens4}$ | IncRes-v2$_{ens}$ |
|---|---|---|---|---|---|---|---|---|
| Inc-v3 | I-FGSM | **99.9*** | 22.6 | 20.2 | 18.1 | 7.2 | 7.6 | 4.1 |
| | MI-FGSM | **99.9*** | 48.1 | 47.1 | 39.9 | 15.2 | 14.2 | 7.2 |
| | DIM | 99.2* | 69.6 | 64.8 | 58.8 | 22.7 | 21.2 | 10.3 |
| | RT-MI-FGSM(**Ours**) | 96.8* | **71.4** | **68.1** | **62.9** | **30.8** | **28.3** | **14.6** |
| Inc-v4 | I-FGSM | 37.9 | **99.9*** | 26.2 | 21.9 | 8.7 | 8.0 | 5.0 |
| | MI-FGSM | 63.9 | **99.9*** | 53.7 | 47.7 | 19.7 | 16.9 | 9.4 |
| | DIM | 80.1 | 99.0* | 71.4 | 63.6 | 26.6 | 24.9 | 13.4 |
| | RT-MI-FGSM(**Ours**) | **80.4** | 98.5* | **72.5** | **69.0** | **42.6** | **39.1** | **23.4** |
| IncRes-v2 | I-FGSM | 37.2 | 31.8 | **99.6*** | 25.9 | 8.9 | 7.5 | 4.9 |
| | MI-FGSM | 68.6 | 61.9 | **99.6*** | 52.1 | 25.1 | 20.2 | 14.4 |
| | DIM | 80.6 | **76.5** | 98.0* | 69.7 | 36.6 | 32.4 | 22.6 |
| | RT-MI-FGSM(**Ours**) | **80.9** | 75.3 | 96.4* | **71.3** | **48.0** | **41.6** | **33.4** |
| Res-101 | I-FGSM | 27.7 | 23.3 | 21.3 | 98.2* | 9.3 | 7.9 | 5.6 |
| | MI-FGSM | 52.4 | 48.2 | 45.6 | 98.2* | 22.3 | 18.6 | 11.8 |
| | DIM | **71.0** | **65.1** | **62.6** | 97.5* | 32.4 | 29.8 | 17.9 |
| | RT-MI-FGSM(**Ours**) | 66.5 | 61.8 | 59.7 | 96.7* | **33.7** | **29.9** | **20.3** |

We then compare the attack success rates of the RT-MI-FGSM and DIM methods based on data augmentation. The results show that our method mostly performs better on both normally and adversarially trained networks, and RT-MI-FGSM has higher black-box attack success rates than DIM. In particular, compared to DIM, RT-MI-FGSM significantly improves the black-box attack success rates on the adversarially trained networks. For example, when generating adversarial examples on the Inc-v3 network to attack the adversarially trained network Inc-v3$_{ens4}$, the black-box attack success rate of DIM was 16.6%, and that of RT-MI-FGSM was 29.1%. If adversarial examples are crafted on Inc-v4, then RT-MI-FGSM has success rates of 42.6% on Inc-v3$_{ens3}$, 39.1% on Inc-v3$_{ens4}$, and 23.4% on IncRes-v2$_{ens}$, while DIM only obtains corresponding success rates of 26.6%, 24.9%, and 13.4%, respectively.

The results in Table 2 show that RT-DIM, which integrates RT-MI-FGSM and DIM, further improves the attack success rates in mostly black-box settings. For example, when generating adversarial examples on the Inc-v3 network to attack adversarially trained networks, the average attack success rate of RT-DIM reaches 45.5%, while that of the DIM method under the same conditions is 21.6%. The average attack success rate more than doubled with RT-DIM. Interestingly, the white-box attack success rates of RT-DIM are not as high as those of DIM, perhaps because the integration of the two methods further increases the transformation randomness of the original input image. More analysis and discussion about this appears in section 4.4.

TABLE 2: The success rates (%) of adversarial attacks against seven models under single-model setting. Adversarial examples are crafted on Inc-v3, Inc-v4, IncRes-v2, and Res-101, respectively, using DIM and RT-DIM. * indicates white-box attacks.

| Model | Attack | Inc-v3 | Inc-v4 | IncRes-v2 | Res-101 | Inc-v3$_{ens3}$ | Inc-v3$_{ens4}$ | IncRes-v2$_{ens}$ |
|---|---|---|---|---|---|---|---|---|
| Inc-v3 | DIM | **99.2*** | 69.6 | 64.8 | 58.8 | 22.7 | 21.2 | 10.3 |
| | RT-DIM(**Ours**) | 93.9* | **75.3** | **72.0** | **70.4** | **38.9** | **35.5** | **19.0** |
| Inc-v4 | DIM | 80.1 | **99.0*** | 71.4 | 63.6 | 26.6 | 24.9 | 13.4 |
| | RT-DIM(**Ours**) | **85.7** | 96.1* | **77.5** | **74.9** | **54.0** | **48.3** | **34.1** |
| IncRes-v2 | DIM | **80.6** | **76.5** | **98.0*** | 69.7 | 36.6 | 32.4 | 22.6 |
| | RT-DIM(**Ours**) | 77.6 | 73.4 | 89.6* | **70.0** | **51.1** | **45.8** | **38.0** |
| Res-101 | DIM | **71.0** | 65.1 | 62.6 | **97.5*** | 32.4 | 29.8 | 17.9 |
| | RT-DIM(**Ours**) | 70.8 | **65.5** | **62.9** | 94.0* | **40.2** | **36.8** | **25.2** |

*4.3 Attacking an ensemble of networks*

Though RT-MI-FGSM and RT-DIM can improve the transferability of adversarial examples on the black-box models, we can further increase their attack success rates by attacking the ensemble models. We follow the strategy in [9] to attack multiple networks simultaneously. We consider all seven networks discussed above. Adversarial examples are crafted on an ensemble of six networks, and tested on the ensembled network and hold-out network, using I-FGSM, MI-FGSM, DIM, RT-MI-FGSM, and RT-DIM, respectively. The number of iterations in the iterative method is $T = 10$, the perturbation size is $\varepsilon = 16$, and the ensemble weights of networks are equal, *i.e.*, $\omega_k = 1/6$.

The experimental results are summarized in Table 3, which shows that in the black-box settings, RT-DIM has higher attack success rates than the other methods. For example, with Inc-v3 as a hold-out network, the success rate of RT-DIM attacking Inc-v3 is 85.2%, while those of I-FGSM, MI-FGSM, DIM, and RT-MI-FGSM are 54.3%, 75.4%, 83.7%, and 84.3%, respectively. On challenging adversarially trained networks, the average success rate of RT-DIM for black-box attacks is 72.1%, which is 23.5% higher than that of DIM. These results show the effectiveness and advantages of our method.

TABLE 3: The success rates (%) of adversarial attacks against seven models under multi-model setting. The "-" symbol indicates the name of the hold-out network. Adversarial examples are generated on the ensemble of the other six networks. The first row shows success rates for the ensembled networks (white-box attack), and the second row shows success rates for the hold-out network (black-box attack).

| | Attack | -Inc-v3 | -Inc-v4 | -IncRes-v2 | -Res-101 | -Inc-v3$_{ens3}$ | -Inc-v3$_{ens4}$ | -IncRes-v2$_{ens}$ |
|---|---|---|---|---|---|---|---|---|
| Ensemble | I-FGSM | 98.5 | 98.2 | 99.1 | **97.8** | 98.2 | 98.2 | 96.3 |
| | MI-FGSM | **98.8** | **98.7** | **99.4** | **97.8** | **98.6** | **98.5** | **96.8** |
| | RT-MI-FGSM(**Ours**) | 93.9 | 95.4 | 96.2 | 93.5 | 95.5 | 95.0 | 96.3 |
| | DIM | 87.2 | 86.8 | 87.7 | 87.8 | 91.9 | 91.4 | 91.8 |
| | RT-DIM(**Ours**) | 87.7 | 87.8 | 90.0 | 89.0 | 91.3 | 92.8 | 91.9 |
| Hold-out | I-FGSM | 54.3 | 48.3 | 48.8 | 41.2 | 17.4 | 18.5 | 10.8 |
| | MI-FGSM | 75.4 | 69.7 | 67.5 | 62.8 | 25.4 | 31.2 | 19.1 |
| | RT-MI-FGSM(**Ours**) | 84.3 | 80.9 | 80.7 | 77.1 | 54.0 | 57.8 | 40.7 |
| | DIM | 83.7 | 82.3 | 80.5 | 76.9 | 49.8 | 52.0 | 41.7 |
| | RT-DIM(**Ours**) | **85.2** | **82.4** | **84.0** | **81.9** | **75.2** | **75.7** | **65.4** |

In the white-box settings, we encounter a similar result to that of RT-MI-FGSM and RT-DIM above (see Section 4.2): the white-box attack success rates of RT-MI-FGSM on the ensemble models are lower than those of MI-FGSM, but are higher than those of RT-DIM and DIM, and the results of RT-DIM are lower than those of DIM and RT-MI-FGSM. This is an interesting result. Perhaps the model and method ensembles have something in common, which have similar effects on the generation of adversarial examples. It remains an open issue for future research.

*4.4 Hyper-parameter studies*

In this section, we conduct extended experiments to further study the influence of different parameters on RT-MI-FGSM and RT-DIM. We consider attacking an ensemble of networks to more accurately evaluate the robustness of the models [14]. The experimental settings are maximum perturbation $\varepsilon = 16$, number of iterations $T = 10$, and step size $\alpha = 1.6$. For MI-FGSM, the decay factor is defaulted to $\mu = 1.0$, and we use default settings for DIM [14].

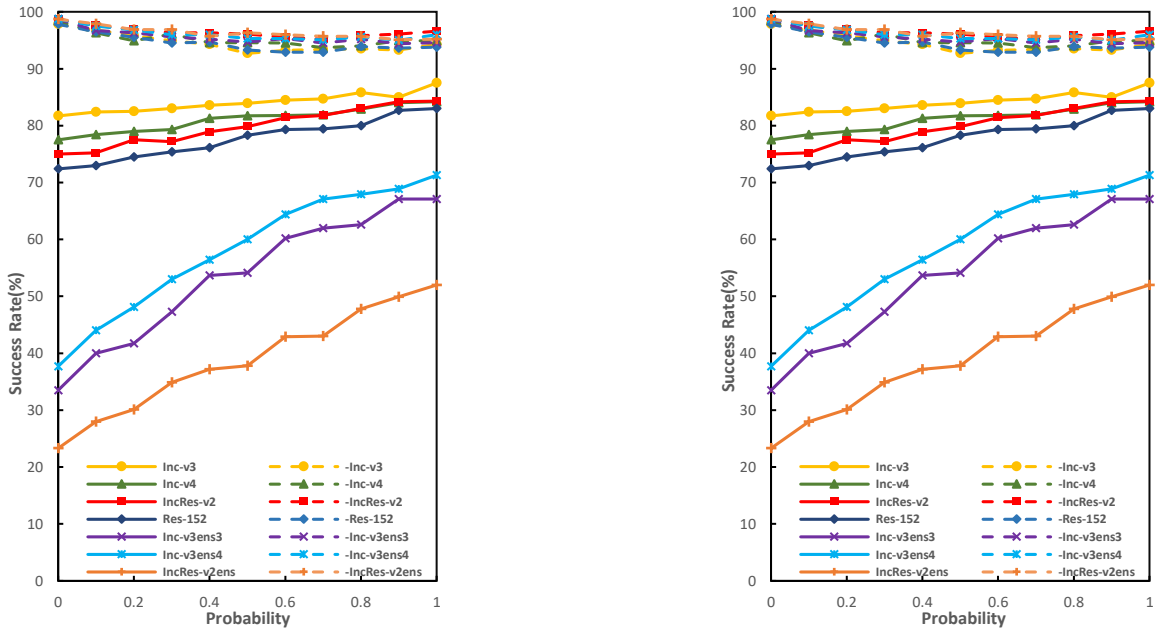

FIGURE 1: Success rates of RT-MI-FGSM (*left*) and RT-DIM (*right*) under different transformation probabilities $p$. Adversarial examples are generated on an ensemble of six networks, and tested on the ensembled network (*white-box setting, dashed line*) and hold-out network (*black-box setting, solid line*).

*Transformation probability* $p$. We first study the impact on the attack success rates in the white-box and black-box models of varying $p$ between 0 and 1. When $p = 0$, RT-MI-FGSM degrades to MI-FGSM, and RT-DIM to DIM. Figure 1 shows the attack success rates of our method on various networks. We can see that the trends of RT-MI-FGSM and RT-DIM are different with the increase of $p$. For RT-MI-FGSM, as $p$ increases, the success rates for black-box attacks increase and that for white-box attacks decrease. For RT-DIM, with the increase of $p$, the success rates of black box attack on adversarially trained networks increase steadily. The success rates of white-box attacks and black-box attacks on normally trained network first increase and then decrease, and finally present an upward trend. Moreover, for all attacks, if $p$ is relatively small, *i.e.*, only a small number of randomly transformed inputs are utilized, the black-box success rates on adversarially trained models increase significantly, the black-box success rates on normally trained networks fluctuate slightly, while the white-box success rates only drop a little. This result shows the importance of adding randomly transformed inputs to the generation of adversarial examples.

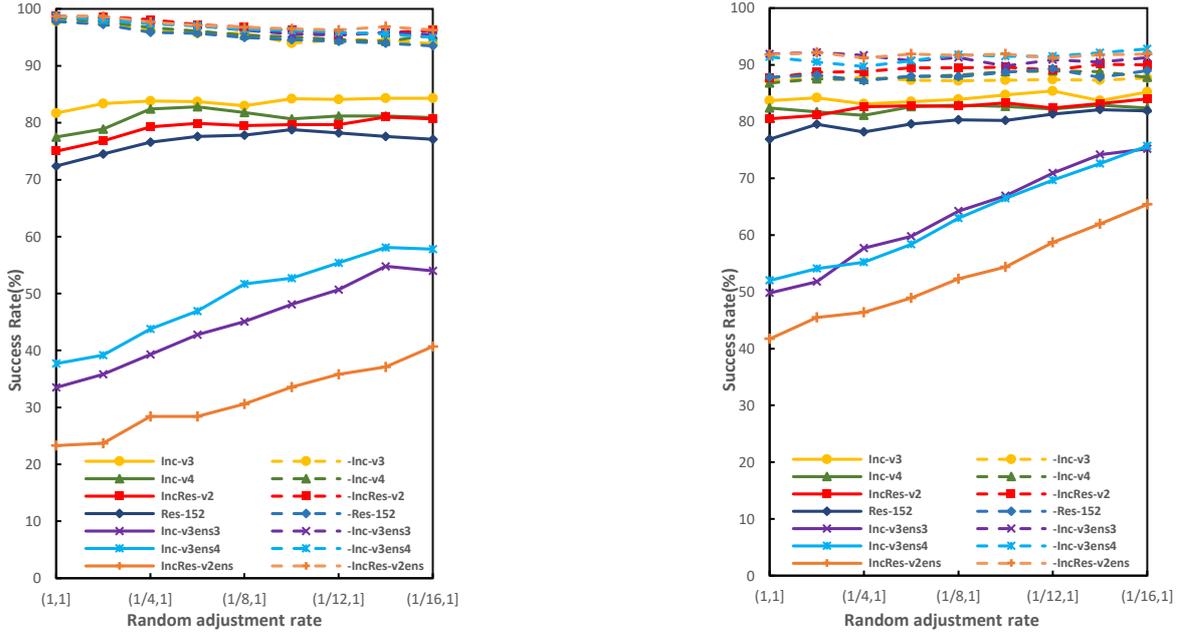

FIGURE 2: Success rates of RT-MI-FGSM (*left*) and RT-DIM (*right*) under different random adjustment rates $r$. Adversarial examples are generated on an ensemble of six networks and tested on the ensembled network (*white-box setting, dashed line*) and hold-out network (*black-box setting, solid line*).

*Random adjustment rate* $r$. We study the effect of $r$ on the attack success rates under the white-box and black-box settings, randomly selecting $r$ within a range, which is changed from $(1,1]$ to $(1/16,1]$. When $r \in (1,1]$, *i.e.*, $r=1$, RT-MI-FGSM degrades to MI-FGSM, and RT-DIM degrades to DIM. The attack success rates on various networks are shown in Figure 2. It can be seen that as the value range of $r$ increases, the success rates for black-box attacks of RT-MI-FGSM increase, and they decrease for white-box attacks. However, for RT-DIM, the success rates of black-box attacks on adversarially trained networks are significantly improved, and the success rates of white-box attack and black-box attacks on normally trained networks are slightly increased. The random adjustment rate $r$ considers both the randomness of the image brightness transformation and adjustment amplitudes. We next adjust the amplitudes of images separately, and discuss the constant adjustment rate $r$.

*Constant adjustment rate* $r$. We finally study the influence of a constant value of $r$ on the attack success rates under the white-box and black-box settings, changing the value range of $r$ from $1/16$ to $1$, *i.e.*, the value of $r$ is increasing. In each iteration, the transformation of the original input is the same. When $r=1$, RT-MI-FGSM degrades to MI-FGSM, and RT-DIM to DIM. Figure 3 shows the attack success rates on various networks. It should be noted that the leftmost ordinate value in Figure 3 represents the attack success rate of our method (RT-MI-FGSM and RT-DIM), while the ordinate value at the far right of Figure 1 and Figure 2 represents the attack success rate of our method. As $r$ increases, the attack success rates of RT-MI-FGSM and RT-DIM have different trends. For RT-MI-FGSM, with the increase of $r$, *i.e.*, the amplitude of image brightness transformation decreases, the success rates of black box attacks on adversarially trained networks decrease significantly. The success rates of black box attacks on normally trained networks first increase and then decrease, and finally present a downward trend, while the white-box attack success rates increase slightly. For RT-DIM, the success rates for black-box attacks on the adversarially trained networks decrease greatly, while they drop a little for white-box and black-box attacks on the normally trained networks. We also found that the same methods have different attack effects under random and constant adjustment rates. With a random adjustment rate, a higher attack success rate is more easily achieved in the white-box models. With a constant adjustment rate, it is easier to obtain a higher attack success rate on the normally trained networks and in the black-box models. These results provide useful suggestions for constructing strong adversarial attacks in practice.

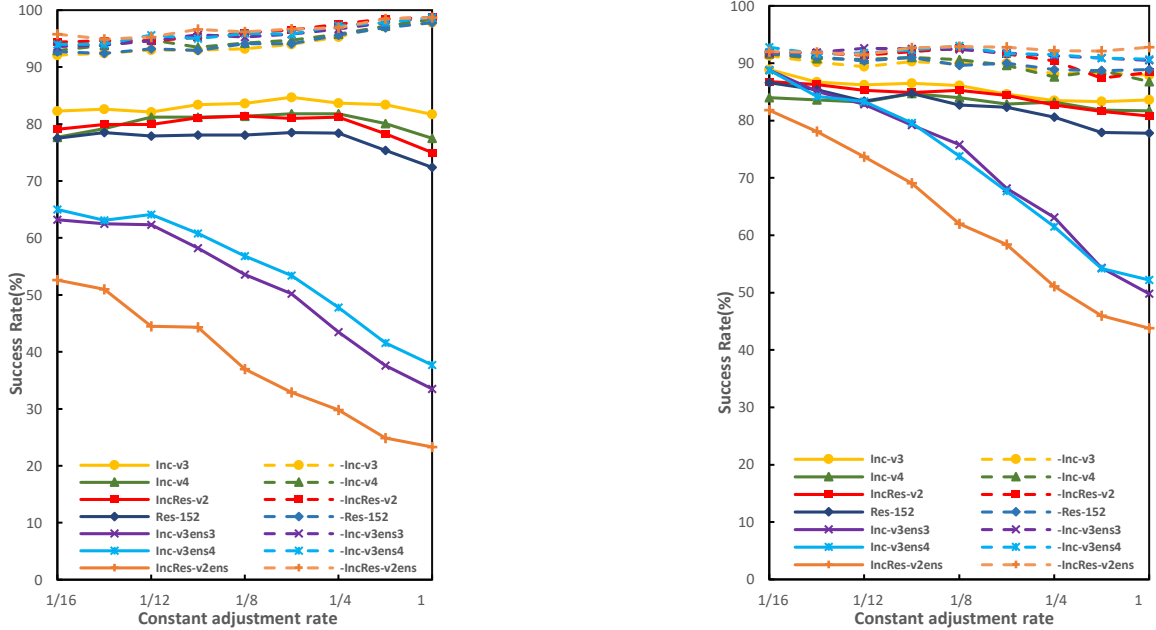

FIGURE 3: Success rates of RT-MI-FGSM (*left*) and RT-DIM (*right*) under different constant adjustment rates $r$. Adversarial examples are generated on an ensemble of six networks and tested on the ensembled network (*white-box setting, dashed line*) and hold-out network (*black-box setting, solid line*).

*4.5 Discussion*

A novel viewpoint [9] indicates that gradient-based adversarial example generation is similar to the training process of DNNs, and the transferability of adversarial examples is similar to the generalization of models, hence some methods used to improve model generalization performance in the training process can be utilized in the generation of adversarial examples to improve their transferability. Inspired by data augmentation [1,2], we perform random brightness transformation of the input image to effectively alleviate overfitting and improve the transferability of adversarial examples. Moreover, we test other data augmentation methods, such as random erasing and random addition of Gaussian noise, but they have little effect on the transferability of adversarial examples. This inspires us to continue to explore the nature of adversarial examples, study the differences between data augmentation methods, and explore more ways to improve model generalization performance. Then adversarial examples with better transferability can be generated to evaluate and improve the robustness of models.

## 5.Conclusions

In this paper, we propose a new attack method based on data augmentation that randomly transforms the brightness of the input image at each iteration in the attack process to alleviate overfitting and generate adversarial examples with more transferability. Compared with traditional FGSM related methods, the results on the ImageNet dataset show that our proposed attack method has much higher success rates for black-box models, and maintains similar success rates for white-box models. In particular, we combine our method with DIM to form RT-DIM to further improve the success rates for black-box attacks on adversarially trained networks. Moreover, we use the method of attacking ensemble models simultaneously to further improve the transferability of adversarial examples. The results of this enhanced attack show that the average black-box attack success rate of RT-DIM on adversarially trained networks outperforms DIM by a large margin of 23.5%. Our work of RT-MI-FGSM suggests that other data augmentation methods may also be helpful to build strong attacks, which will be our future work, and the key is how to find effective data augmentation methods for iterative attacks. Our proposed attack method can help evaluate the robustness of the models and the effectiveness of different defense methods and build deep learning models with higher security.

## Appendix A. Details of Algorithms

The algorithm of RT-DIM attack is summarized in Algorithm 2. We can obtain the RT-MI-FGSM attack algorithm by removing step 4 of Algorithm 2 and obtain the DIM attack algorithm by removing step 5. In addition, the MI-FGSM attack algorithm can be obtained by removing steps 4 and 5. Of course, our method can also be related to the family of Fast Gradient Sign Methods by adjusting the transformation probability $p$ and decay rate $\mu$.

---

**Input:** A clean example $x$ with ground-truth label $y$; a classifier $f$ with loss function $J$;
**Input:** Perturbation size $\varepsilon$; maximum iterations $T$ and decay factor $\mu$.
**Output:** An adversarial example $x^{adv}$

1: $\alpha = \varepsilon / T$
2: $x_0^{adv} = x$; $g_0 = 0$
3: **for** $t = 0$ to $T-1$ **do**
4:   Get $x_t^{adv}$ by $x_t^{adv} = RT(x_t^{adv}; p)$ ▷ apply random transformation of the input's brightness with the probability $p$
5:   Get $x_t^{adv}$ by $x_t^{adv} = T(x_t^{adv}; p)$ ▷ apply random resizing and padding to the inputs with the probability $p$
6:   Get the gradients by $\nabla_x J(\theta, x_t^{adv}, y)$
7:   Update $g_{t+1}$ by $g_{t+1} = \mu \cdot g_t + \dfrac{g}{\| g \|_1}$
8:   Update $x_{t+1}^{adv}$ by Eq. (5)
9: **return** $x^{adv} = x_T^{adv}$

---

ALGORITHM 2: the details of RT-DIM

## Appendix B. Visualization of Images

The six randomly selected original images and corresponding randomly transformed images and generated adversarial samples are shown in Figure 4. The adversarial examples are crafted on Inc-v3 by the RT-DIM method.

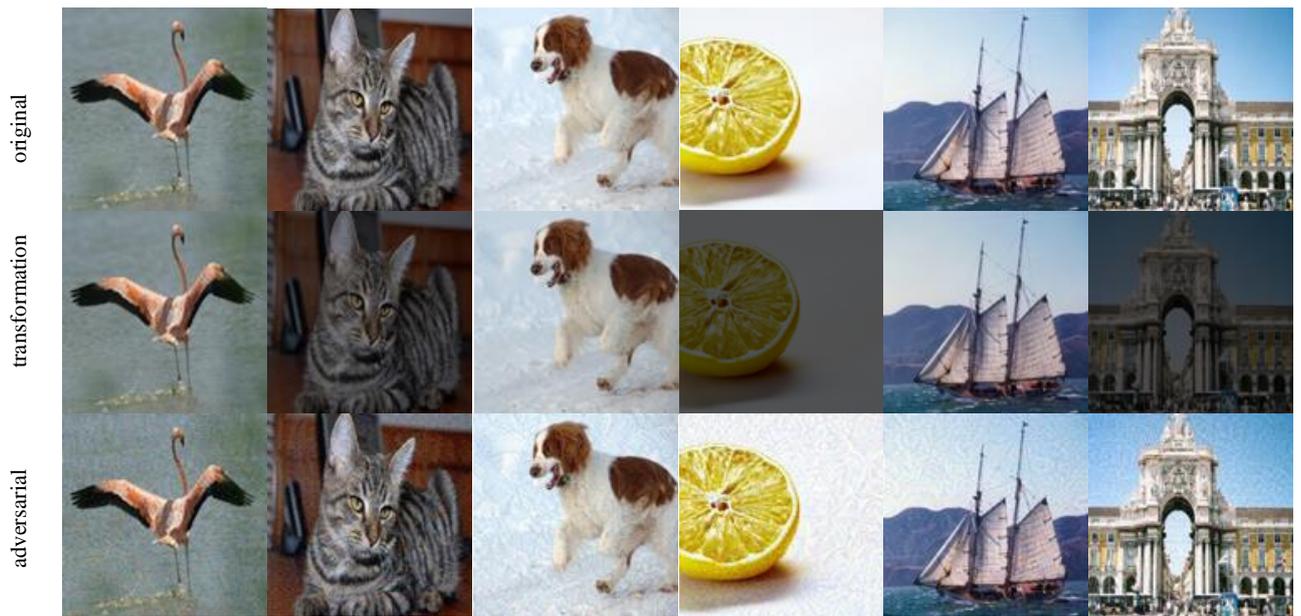

FIGURE 4: Images from first to third line are original inputs, randomly transformed images, and generated adversarial examples, respectively.